\documentclass[conference]{IEEEtran}
\usepackage{epsfig}
\usepackage{graphicx}

\ifCLASSINFOpdf
\else
\fi
\begin{document}
\title{Motion Saliency Based Automatic Delineation of Glottis Contour in High-speed Digital Images}

\author{\IEEEauthorblockN{Xin Chen}
\IEEEauthorblockA{Emerging Technology Center\\
Midea Group\\
San Jose, CA, 95134, USA\\
Email: chen1.xin@midea.com}
\and
\IEEEauthorblockN{Emma Marriott}
\IEEEauthorblockA{Department of Computer Science\\
Stanford University\\
Stanford, CA 94305, USA\\
Email: emarriott@stanford.edu}
\and
\IEEEauthorblockN{Yuling Yan}
\IEEEauthorblockA{Department of Department of Bioengineering\\
University of Santa Clara\\
Santa Clara, CA 95053, USA\\
Email: yyan1@scu.edu}}


%Telephone: (800) 555--1212, Fax: (888) 555--1212}


\maketitle

\begin{abstract}
In recent years, high-speed videoendoscopy~(HSV) has significantly aided the diagnosis of voice pathologies and furthered the understanding the voice production in recent years. As the first step of these studies, automatic segmentation of glottal images till presents a major challenge for this technique. In this paper, we propose an improved Saliency Network that automatically delineates the contour of the glottis from HSV image sequences. Our proposed additional saliency measure, Motion Saliency~(MS), improves upon the original Saliency Network by using the velocities of defined edges. In our results and analysis, we demonstrate the effectiveness of our approach and discuss its potential applications for computer-aided assessment of voice pathologies and understanding voice production.
\end{abstract}


% \fi
\IEEEpeerreviewmaketitle

\section{INTRODUCTION}
Nowadays, high-speed videoendoscopy (HSV) has become more common in assessing voice condition and understanding mechanism of pronation~\cite{c1}\cite{c2}\cite{c3}. In addition to reduce cost of HSV systems, this popularization can be attributed to two major developments. First, HSV systems are now capable of recoding speed at 2,000 - 4000 frames per second~(fps)-a speed fast enough to resolve the actual vibrations of vocal folds. Second, the resolution of each frame has increased to 512$\times$512 or higher, which allows each frame to preserve  the real shape of subtle parts of the glottis contours.

In recent years, clinicians and speech scientists have developed numerous techniques to analyze HSV imaging data with a focus on the glottis~\cite{c4}\cite{c5}\cite{c6}\cite{c7}. In many of these techniques, automatic segmentation and delineation of the contours of glottis are both a critical and bottleneck step for successful analysis.

\begin{figure}[!tb]
\begin{center}
 %\includegraphics[width=0.8\linewidth]{lowmode.eps}
 \includegraphics[width=0.8\linewidth]{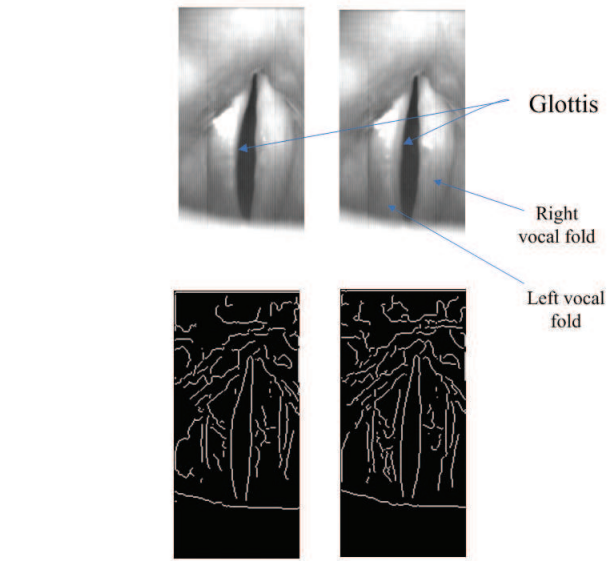}
\end{center}
  \caption{Top: two adjacent frames of laryngeal images from an HSV imaging recording. The air space between the vocal folds, or the opening region, is glottis. The normal vocal fold motion in a sustained vowel phonation consists of open-close cycles. Bottom: Two edge images corresponding to top row images with Canny edge operator. As shown in top row, the image quality of the frames in HSV image is noisy. Most of the image frames don't contain rich texture information. In the bottom row, the edges are achieved by Canny edge operator. It is clear that contours of glottis in each image frame of HSV recording are salient curves.}
\label{fig:1}
\end{figure}

As shown in Figure~\ref{fig:1}, poor image quality and false or ambiguous edges complicate the task of segmentation of vocal folds from HSV images. Although some efforts have been made, there is not a perfect solution till now. Some of the proposed involve applying active contour model (Snake)~\cite{c00}, a well-known energy-minimization algorithm, to delineate the contour of glottis~\cite{c8}\cite{c9}. The common disadvantage of these methods is that the active contour model fails to delineate sharp concerns, such as top portion and bottom of the glottis. X. Chen et al. introduced a three-step region-growing method to segment glottal area~\cite{c10}. However, the approach depended heavily on an accurate and reliable of a seed-pixel point. A. Mendez et. al employed Gabor filters to segment glottal area~\cite{c11}. Although this approach is well-suited for images with rich texture information. HSV images frequently lack rich texture structures, as shown in Figure~\ref{fig:1}.

In this paper, we propose an improved Saliency Network to automatically delineate contour of glottis from HSV data. Our approach is based on original work on The Saliency Network introduced by Saashua and Ullman~\cite{c12}, which is considered as one of the top methods to solve the problem of extracting salient images with gap completion. Saliency measure of the Saliency Network, referred to as SU measure, is formulated from both local saliency and structural saliency. The SU measure, however, is poor in computing structural saliency because it lacks enough properties~\cite{c13}, which sometimes leads to failure in finding real contours.

To address this issue, we propose a novel motion-derived saliency measure, called motion saliency (\textbf{MS}) measure. Since motion contains rich visual information on image structure and shape, our proposed MS saliency measure can effectively represent the missing structural saliency of the current Saliency Network.

This paper makes two major contributions as following:
\begin{enumerate}
  \item We propose a novel MS measure that formulates from motion information in HSV imaging sequences. Combining with the MS measure, the Saliency network has greatly improved in structural saliency representation;
  \item The improved Saliency Network has been successfully applied in automatical delineation of glottis contours from high-speed laryngeal images. These results have demonstrated its potential applications in the subsequent computer-aided assessment of voice pathologies and understanding voice production;

\end{enumerate}

\section{PROPOSED METHOD}\label{method}

\subsection{Background of the Saliency Network}
The Saliency Network aims at extracting salient contour from images and has been considered as one of the top bottom-up methods of perceptual group. In the Saliency Network, image pixels, also called elements~\cite{c12}, are classified into active and virtual elements. The elements that lie on edges are referred to as active elements, otherwise they are referred as to virtual elements.

Given a curve $\Gamma$ composing of the $(N+1)$ elements, the SU measure of $\Gamma$ is defined as:
\begin{equation}
\label{eq:1_1}
\Phi_{su}(\Gamma) = \sum\limits_{j=i}^{i+N} C_{i,j}\rho_{i,j}\sigma_{j}
\end{equation}
where $\sigma_{j}$ is set to $1$ for an active element, otherwise it is set to $0$. The attenuation function $\rho_{i,j}$ is defined as:
\begin{equation}
\label{eq:1_2}
\rho_{i,j} = \prod\limits_{k=i+1}^{j} \rho_{k}
\end{equation}
where $\rho_{i,i} = 1$, and $\rho$ is in the range $(0, 1]$. If $\rho_{i}$ is active it is set to as a value smaller than or equal to $1$, and if $\rho_{i}$ is virtual it is set as smaller than the value of the active. In ~\cite{c12}, $\rho$ is set to $1$ if $\rho_{i}$ is active, otherwise it is set to $0.7$. $C_{i,j}$ aims to measure the shape of the curve that is inversely related to the total curvature of the curve and is defines as:
\begin{equation}
\label{eq:1_3}
C_{i,j} = e^{-\int_{p_{i}}^{p_{j}} (\frac{d\theta}{ds})^{2}ds}
\end{equation}
where $\theta(s)$ is the slope along the curve $s$.

The SU measure defined in Equation~\ref{eq:1_1} is a weighted contribution to the local saliency values $\sigma_{j}$ along the curve. The weight consists of two factors, $C_{i,j}$ and $\rho_{i,j}$. The first inversely represents the number of virtual elements of the curve, and the second inversely represents the total curvature of the curve.

The saliency of an element $p_{i}$ is defined as the maximum of all curve starting from $p_{i}$:
\begin{equation}
\label{eq:1_4}
\Phi(i) =\max\limits_{\Gamma\in\Omega(i)}\Phi_{su}(\Gamma)
\end{equation}
where $\Omega(i)$ denotes the set of curves starting from $p_{i}$.

\subsection{Motion Saliency Measure}
Since motion is an effective cue to represent image structure~\cite{c14}, MS measure calculated from motion cue effectively represents the structural saliency.

The formulation of the MS measure is based on different kinetic and kinematic properties corresponding to different types of motion. Since the image acquisition rate of HSV is some 10 times of the fundamental frequency of the vocal fold vibration, it is reasonable to be considered as a case of non-rigid object with low-order model shapes and slow variation load. As a matter of fact, most cases in medical images belong to this type of case of applications.

Given this type of motion pattern, the proposed MS measure is formulated based on the following two motion features:
\begin{itemize}
\item It has the maximum sum of the magnitude of velocity of all elements;
\item The velocity of the adjacent elements of the contour are the closest. In other words, given any adjacent elements of the contour, both angular and spatial distances of the two vectors are closest;
\end{itemize}

\textbf{Velocity Estimation} The velocity is estimated from Lucas Kanade (\textbf{LK}) algorithm that is a widely used optical flow method~\cite{c15}. While optical flow method is used to estimate rigid motion, the velocity values are reliable for non-rigid motion in some special applications, such as our work, for following two reasons:
\begin{itemize}
\item The motion of small patches between adjacent frames reasonably approximates to rigid motion.
\item The observed brightness of any object point is constant over time;
\end{itemize}
As a result, the computation of MS measure based on these velocity estimations should also be reliable as well.

\textbf{Normalization} In order to smooth and normalize the velocity image estimated from LK algorithm, we take hyperbolic tangent function as a nonlinear transformation of each velocity image. The function is defines as:
\begin{equation}
\label{eq:2_5}
f(x) = \frac{e^{\lambda_{1}x} - e^{-\lambda_{1}x}}{e^{\lambda_{1}x} + e^{-\lambda_{1}x}}
\end{equation}
where $\lambda_{1}$ is a positive constant. In our work, we set $\lambda_{1}$ to $0.5$.

\textbf{Angular Distance} Given the normalized velocity $\overrightarrow{V} = \{V_{x}(i), V_{y}(i)\}$ at the point $i$ of the curve, the angular distance between point $i$ and point $j$ is the cosine of included angle of these two vectors, and is defined as:
\begin{equation}
\label{eq:2_2}
D_{a}(i,j) = \frac{V_{x}(i)\times V_{x}(j) + V_{y}(i)\times V_{y}(j)}{\sqrt{V_{x}^{2}(i) + V_{y}^2(i)}\sqrt{V_{x}^{2}(j) + V_{y}^{2}(j)}}
\end{equation}
where the value of $D_{a}$ is in the range of $[-1, 1]$. A unity value of the angular distance implies that the two vectors are most-collinea0r or have maximum  similarity, and -1 implies the opposite.

\textbf{Spatial Distance} Spatial distance aims to measure distance between two vectors in space, and is defined as
\begin{equation}
\label{eq:2_3}
D_{s}(i,j) = \frac{2}{e^{-\lambda_{2}D(i,j)} + e^{\lambda_{2}D(i,j)}}
\end{equation}

\begin{equation}
\label{eq:2_4}
D(i,j) = \sqrt{(|V_{x}(i)| - |V_{x}(j)|)^{2} + (|V_{y}(i)| - |V_{y}(j)|)^2}
\end{equation}
where $\lambda_{2}$ is a positive constant. $D(i,j)$ is the Euclidean distance of two vector. The value of spatial distance is a positive constant in the range of $(0,1]$. The $\lambda_{2}$ is set to $0.5$ in our work.

\textbf{MS Measure} On the basis of the angular distance and the spatial distance, given a curve $\Gamma$ composed of the $(N+1)$ elements, we define the MS measure of $\Gamma$ as:
\begin{equation}
\label{eq:2_6}
\Phi_{ms}(\Gamma) = \sum\limits_{j=i}^{i+N} M(i)D_{a}(i,j)D_{s}(i,j)
\end{equation}
Where $M(i)$ is the magnitude of the normalized velocity $\overrightarrow{V(i)}$ at point $i$ with
\begin{equation}
\label{eq:2_10}
M(i) = \sqrt{V_{x}^{2}(i) + V_{y}^{2}(i)}
\end{equation}

The MS measure defined in Equation~\ref{eq:2_6} is a weighted contribution of $M(i)$ along the curve. The weight is a product of angle distance and spatial distance. If the two vectors are same, both the angular distance and the spatial distance are unity.

\subsection{Proposed Improved Saliency Network}
We define the improved Saliency Network as:
\begin{equation}
\label{eq:2_1}
\Phi(\Gamma) = \alpha\Phi_{su}(\Gamma) + \beta\Phi_{ms}(\Gamma)
\end{equation}
where $\alpha$ and $\beta$ are weighted coefficient of SU measure and MS measure, respectively, and $\Phi(\Gamma)$ is a weighted sum of SU and MS measures.

Through the introduction of an additional term, $\Phi_{ms}(\Gamma)$, our method, compared to original Saliency Network $\Phi_{su}(\Gamma)$, improves on computation of structural saliency. The parameters $\alpha$ and $\beta$ are set to the range of $[0$ $1]$ and defined by users according for different cases.

Substituting $\Phi(\Gamma)$ for $\Phi_{su}(\Gamma)$ of the Equation~\ref{eq:1_4}, the saliency of an element $p_{i}$ is rewritten as the maximum of all curve starting from $p_{i}$:
\begin{equation}
\label{eq:2_7}
\Phi(i) =\max\limits_{\Gamma\in\Omega(i)}\Phi(\Gamma)
\end{equation}
where $\Omega(i)$ denotes the set of curves starting from $p_{i}$.

Essentially, the improved Saliency Network presents an optimization problem, and we adopted the dynamic programming is adopted to solve this problem.
\begin{figure}[!tb]
\begin{center}
 %\includegraphics[width=0.8\linewidth]{lowmode.eps}
 \includegraphics[width=0.8\linewidth]{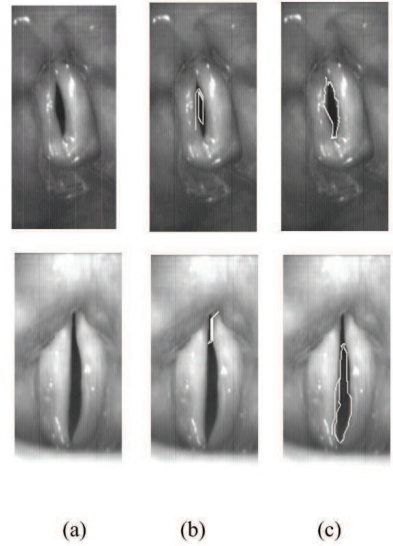}
\end{center}
  \caption{(a): Frames of laryngeal images from an HSV recording. (b): Results of the Saliency Network. (c): Results from our proposed method.}
\label{fig:result}
\end{figure}
\section{RESULTS and ANALYSIS}\label{result}
\subsection{HSV Laryngeal Imaging}
We used a Kay-Pentax (Lincoln, Park, NJ) HSV system to record the laryngeal images. This system is capable of acquiring images at a rate of 4,000 fps with a spatial resolution of 512$\times$512 pixels.

During the examination, clinicians typically insert an endoscope into the subject's open mouth into the posterior pharynx at a ~70-degree angle. To prevent gagging, clinicians took special care in avoiding the posterior pharyngeal wall and velum. The subjects are instructed to produce a sustained vowel /i/ while their tongues were secured. Phonation of the vowel /i/ retracts the epiglottis and base of the tongue, thus opening the laryngeal inlet and providing the best view of the larynx for imaging. The recording time for each HSV recording is typically within 2 seconds, yielding up to 8,000 image frames of data from this procedure.
\subsection{Results and Analysis}
Our experiment adopted a procedure of the subjective evaluation. The major criteria are to evaluate whether the results from glottis contour extraction are sufficiently accurate to facilitate further steps of processing with an ultimate goal of assisting in the computer-aided diagnosis of voice pathologies.

The parameters $\alpha$ and $\beta$ of Equation~\ref{eq:2_1} are set to 0.3 and 0.7, respectively. All of them are empirical values.

The results of analysis obtained from the proposed method as shown in Figure~\ref{fig:result} demonstrated the robustness and accuracy of our method in extracting semantic contour from a noisy background. In comparison to Saliency Network, the computation of the MS measure efficiently represents saliency of structure. The effective motion saliency measure that we introduced enables our method to yield better results than the Saliency Network. As clearly shown in Figure~\ref{fig:result}, our method outperforms the Saliency Network in extracting motion salient contour in noisy scenes from the HSV video.
\section{CONCLUSIONS}\label{conclusion}
We proposed an improved Motion Saliency Network for automatic delineation of glottis contour from the HSV image sequences. The improved Saliency Network consists of two saliency measures, SU measure as defined in~\cite{c12} and MS measure, which is a new motion-derived measure that we introduced in this paper. The MS measure is calculated from motion cues. Since motion efficiently represents global and shape properties of an image, the proposed model better resolves structural saliency properties and should outperform the Saliency Network.

We applied the proposed method to the analysis of sample HSV image sequences and the results have shown that our method was comparable to or better in performance than the existing state-of-the-art methods. Finally, although the improved Saliency Network is applied to glottis image segmentation herein, it is clear that this new model is a universal model because the formulation of the MS measure can vary from different kinetic properties of different types of motion. We anticipate that the proposed model should also be valid for extended applications in medical image analysis, provided reliable velocity estimation is possible.

\section*{ACKNOWLEDGMENT}
The authors would like to thank Dr. Krzysztof Izdebski for his assistance in acquiring HSV data, some of which are used to test the validity of the proposed method.

\end{document}